\documentclass[letterpaper, 10 pt, conference]{ieeeconf}  
\IEEEoverridecommandlockouts                        
\usepackage[T1]{fontenc}    
\usepackage{graphicx} 
\graphicspath{{figures/}} 
\usepackage[bookmarks=true]{hyperref}
\usepackage{amsmath} 
\usepackage{amssymb}  
\usepackage{siunitx}
\usepackage{mathtools}
\usepackage{siunitx}
\usepackage{caption}
\usepackage{subcaption}
\usepackage{booktabs}
\usepackage[utf8]{inputenc}
\usepackage[capitalise]{cleveref}
\usepackage{pgfplots}
\usepackage{lipsum} 
\DeclareUnicodeCharacter{2212}{−}
\usepackage{tikz}
\usepgfplotslibrary{groupplots,dateplot}
\usetikzlibrary{patterns,shapes.arrows, arrows, positioning}
\pgfplotsset{compat=newest, every tick label/.append style={font=\footnotesize}}
\usepackage{pgfkeys}
    \newenvironment{customlegend}[1][]{
        \begingroup
        \csname pgfplots@init@cleared@structures\endcsname
        \pgfplotsset{#1}
    }{
        \csname pgfplots@createlegend\endcsname
        \endgroup
    }
    \def\addlegendimage{\csname pgfplots@addlegendimage\endcsname}

\usepackage{stackengine}
\newcommand\solidcirc[4][0]{\rotatebox{#1}{\tikz{\draw[line width=#2] (0,0) 
  arc [x radius=#3,y radius=#4,start angle=0,end angle=360];}}}

\usepackage[style=ieee, url=false, doi=false, natbib=true, mincitenames=1, maxcitenames=1]{biblatex}
\usepackage[keeplastbox]{flushend}

\title{\LARGE \bf
Double-Prong ConvLSTM for Spatiotemporal Occupancy \\ Prediction in Dynamic Environments
}
 
\author{Maneekwan Toyungyernsub, Masha Itkina, Ransalu Senanayake and Mykel J. Kochenderfer
\thanks{The authors are with Stanford University, CA, USA. 
Email: {\tt\footnotesize \{maneekwt, mitkina, ransalu, mykel\}@stanford.edu}.}
}

\begin{document}

\maketitle
\thispagestyle{empty}
\pagestyle{empty}
\tikzstyle{block}=[align=center,fill=gray!10,rounded corners,draw=black]

\begin{abstract}
Predicting the future occupancy state of an environment is important to enable informed decisions for autonomous vehicles. Common challenges in occupancy prediction include vanishing dynamic objects and blurred predictions, especially for long prediction horizons. In this work, we propose a double-prong neural network architecture to predict the spatiotemporal evolution of the occupancy state. One prong is dedicated to predicting how the static environment will be observed by the moving ego vehicle. The other prong predicts how the dynamic objects in the environment will move. Experiments conducted on the real-world Waymo Open Dataset indicate that the fused output of the two prongs is capable of retaining dynamic objects and reducing blurriness in the predictions for longer time horizons than baseline models.
\end{abstract}

\section{INTRODUCTION}
A key component to successful deployment of autonomous vehicles (AVs) is their ability to safely and intelligently maneuver the shared space. Urban driving scenes are particularly challenging for autonomous navigation due to crowded driving environments with many different types of road users. Human drivers can plan ahead and drive safely because they anticipate how their immediate surroundings will evolve based on their current surroundings. Similarly, the capability to accurately predict the temporal evolution of the environment would enable AVs to proactively plan safe trajectories. 

To make predictions, a representation of the local environment around the AV is required, typically in the form of a map.
Due to its simplicity, the occupancy grid map (OGM)~\cite{Elfes} is often used in practice. OGMs can be generated from on-board range-bearing sensor measurements (e.g. from LiDAR or radar) and represent the local environment by discretizing the space into grid cells. Each cell contains the belief of its respective occupancy probability, and is assumed to be independent of all other cells. Sensor readings can be incorporated into the OGM representation using a recursive Bayesian filter~\cite{Elfes}. An alternative approach is to update the grid cells using Dempster--Shafer Theory (DST)~\cite{dst} to produce evidential occupancy grid maps (eOGMs)~\cite{Pagac}. Unlike OGMs, which only consider the binary \textit{free} or \textit{occupied} hypotheses for a grid cell, each cell in an eOGM carries additional information on the \textit{occluded} occupancy hypothesis. This extra information channel allows the eOGMs to distinguish between lack of information (e.g. occlusion) and uncertain information (e.g. moving objects)~\cite{Masha}.   

\begin{figure}[t]
\centering
\input{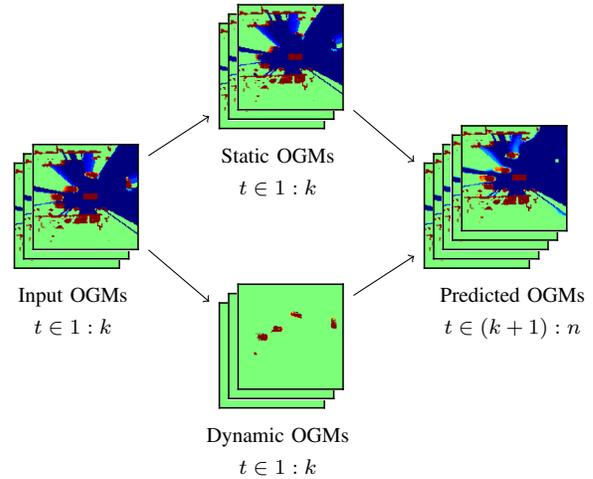}
\caption{\small Future occupancy states are predicted by considering the globally static and moving components separately. 
}
\label{fig:teaser}
\end{figure}

OGMs hold structural similarities with RGB images as they both depict discretized spatial information. Hence, the spatiotemporal task of predicting the future state of the environment can be posed as a video frame prediction task. Consequently, recent work has adapted deep neural network architectures designed for video frame prediction to predict OGMs instead~\cite{Masha, Schreiber, Mohajerin_2019_CVPR}. \citet{Masha} use a convolutional long-short term memory (ConvLSTM)~\cite{convlstm} architecture in the form of the Predictive Coding Network (PredNet)~\cite{PredNet} to predict the environment state represented as eOGMs. The paper also investigates the benefits of incorporating dynamic state information (i.e. velocity) into the eOGMs. This extended representation is known as a dynamic occupancy grid map (DOGMa), where the cell-wise velocities are estimated using a particle filter and carried in an extra channel alongside the occupancy information~\cite{Nuss_DOGMa}. \citet{Masha} find that the prediction performance between the eOGM and DOGMa representations is relatively unchanged, despite DOGMas providing additional velocity information. Furthermore, dynamic objects tend to disappear in longer time horizon predictions.

We extend the work by \citet{Masha} to develop a model that incorporates environment dynamics directly within its architecture, rather than only as a data channel input. Our method aims to increase the longevity of the dynamic objects in the predictions for longer time horizons. We draw inspiration from human perception. Human drivers continually perceive and distinguish between the static and dynamic parts of the environment, and use this information to predict the future local environment around their vehicle. Thus, we design our model to have two sub-architectures that learn the motion models for the moving objects and the globally static environment, as shown in \cref{fig:teaser}. The prediction of the globally static environment captures the relative motion of the static environment with respect to the moving ego vehicle.
We propose identifying the moving components in the environment via an object-centric approach, instead of computing expensive cell-wise velocity estimates to encode the environment dynamics. 

Our contributions in this work are as follows. We develop a double-prong deep neural network architecture that effectively incorporates the environment dynamics to predict future eOGMs. Each prong is dedicated to predicting the future static and dynamic components in the environment, represented as separate eOGMs.
The prong outputs are combined to form the complete predicted eOGMs. We show that our model maintains dynamic objects for longer time horizons and outperforms other state-of-the-art architectures. Moreover, our model is able to handle highly sparse data in the dynamic input, as moving obstacles account for significantly fewer grid cells than static or free space. The proposed framework is evaluated on the real-world Waymo Open Dataset~\cite{Sun_2020_CVPR}.          

\section{RELATED WORK}
\label{related works}
\subsection{Video Frame Prediction}
Deep learning has been studied and used successfully in spatiotemporal prediction problems in computer vision, such as video frame prediction. The PredNet model learns to predict future video frames by propagating the error terms between predictions and targets vertically and laterally within its recurrent ConvLSTM architecture~\cite{PredNet}. The model succeeds at predicting the movement of synthetic objects and natural images from a camera on-board a vehicle. The PredRNN++~\cite{predrnn++} architecture addresses the difficulties in gradient propagation for deep-in-time networks, which have a high number of recurrent states between the input and output. The advantage of deep-in-time networks is the improved learning of spatial correlations and short-term video dynamics~\cite{predrnn++}.
The Memory in Memory (MIM) network~\cite{mim} architecture consists of several LSTM units where the forget gate is replaced by a series of cascaded recurrent modules. The model succeeds at predicting future video frames on both synthetic and real-world datasets. In our proposed environment prediction method, we make use of the PredNet model~\cite{PredNet} following previous work by~\citet{Masha}, and due to its parameter efficient architecture.

\subsection{Double-Prong Architecture}
Double-prong architectures have been used previously for spatiotemporal action recognition and traffic prediction tasks~\cite{Simonyan_two_stream, traffic_model}. \citet{Simonyan_two_stream} develop a double-prong model for action recognition, in which video frames are decomposed into spatial and temporal components. The spatial components consist of still video frames, while the temporal components contain optical flow displacement fields between consecutive frames. Each prong has a deep convolutional neural network (CNN) architecture. Significant improvements in action recognition accuracy are achieved since the temporal inputs explicitly carry motion information, eliminating the need for the model to perform implicit motion estimation.
\citet{traffic_model} present a double-prong CNN model capable of predicting multi-lane traffic speeds. Each prong receives as input the traffic speed and volume data, respectively, and learns to predict traffic flow for each individual lane, outperforming a single-prong alternative. Following the successes of these double-prong models, we consider a double-prong architecture for our proposed environment prediction model.

\subsection{Occupancy Grid Prediction}
In this paper, we study the problem of spatiotemporal environment prediction using OGM representations.
\citet{senanayake2016spatio} propose a theoretical framework for propagating the uncertainty of nearby dynamic objects into the future using reproducing kernel Hilbert space theory under a stationary ego vehicle assumption. \citet{guizilini2019dynamic} extend this model to a moving ego vehicle, but do not predict the local static environment beyond its FOV.
Unlike these methods, our approach does not assume a stationary ego vehicle or keep the stationary areas of the environment fixed, making it more relevant to real-world AV applications.

In contrast to these model-based approaches, several works repurpose deep learning architectures designed for video frame prediction to predict the environment instead. \citet{Schreiber} present an encoder-decoder framework using a ConvLSTM~\cite{convlstm} architecture to predict the future environment represented as DOGMas. They introduce recurrent skip connections into the network to reduce prediction blurriness. However, DOGMas require a particle filter to estimate the cell-wise velocities, which are prohibitively expensive to compute. \citet{Mohajerin_2019_CVPR} introduce a difference-learning-based recurrent architecture to extract and incorporate motion information for OGM prediction. However, unlike our work, they transform the OGMs into a common frame to account for the motion of the ego vehicle. \citet{Masha} study prediction of the future environment using the PredNet~\cite{PredNet} architecture with eOGM and DOGMa data. They find that the PredNet model is able to learn the local motion of the environment sufficiently well from eOGMs without the additional dynamic state information carried in DOGMas. However, dynamic objects still tend to disappear at longer time horizon predictions. Our work extends this approach by intelligently incorporating dynamic information directly into the architecture. We aim to bring direct focus to the dynamic components in the scene without resorting to computing cell-wise velocity estimates. 

\section{APPROACH}
\label{approach}
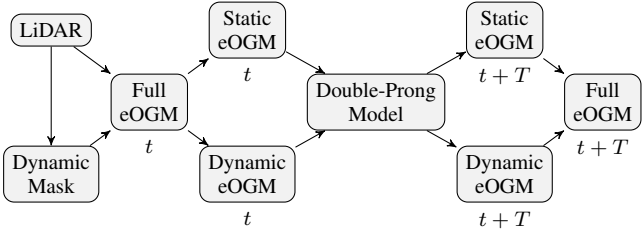
\begin{figure}
    \centering
    \begin{tikzpicture}[>=stealth',font=\footnotesize]
\matrix [column sep=0.15cm,row sep=0.2cm] {
\node[block] (lidar) {LiDAR}; & & \node[block] (staticeogm) {Static\\eOGM}; & & \node[block] (staticeogm2)  {Static\\eOGM}; & \\
& \node[block] (fulleogm1) {Full\\eOGM}; & & \node[block] (double) {Double-Prong \\ Model}; & & \node[block] (fulleogm2) {Full \\ eOGM};  \\
\node[block] (dynamicmasks) {Dynamic \\ Mask}; &  & \node[block] (dynamiceogm) {Dynamic \\ eOGM}; & & \node[block] (dynamiceogm2) {Dynamic \\ eOGM}; & \\
 & & \\
};
\draw [->] (lidar) edge (fulleogm1) edge node[anchor=east,left]{} (dynamicmasks) 
    (fulleogm1) edge (staticeogm)
    (staticeogm) edge (double)
    (dynamiceogm) edge (double)
    (dynamicmasks) edge (fulleogm1)
    (fulleogm1) edge (dynamiceogm)
    (double) edge (staticeogm2)
    (double) edge (dynamiceogm2)
    (staticeogm2) edge (fulleogm2)
    (dynamiceogm2) edge (fulleogm2)
;

\node [below] at (fulleogm1.south) {$t$};
\node [below] at (staticeogm.south) {$t$};
\node [below] at (dynamiceogm.south) {$t$};
\node [below] at (fulleogm2.south) {$t+T$};
\node [below] at (staticeogm2.south) {$t+T$};
\node [below] at (dynamiceogm2.south) {$t+T$};
\end{tikzpicture}
    \caption{\small The pipeline for the proposed framework. The future static and dynamic eOGMs are predicted using the double-prong model, and then fused to form the complete eOGM prediction.}
    \vspace{-0.25cm}
    \label{fig:pipeline}
\end{figure}

This section describes our proposed framework for environment prediction, illustrated in \cref{fig:pipeline}. We form our environment representation using range-bearing sensors (e.g. LiDAR). We convert the point cloud data into eOGMs after filtering out the points belonging to the ground using a Markov Random Field (MRF)~\cite{Postica}, as done by~\citet{Masha}. In order to extract the static and dynamic portions of the environment, we identify moving objects from the filtered point cloud data using a simple thresholding approach. We then generate dynamic masks to construct the dynamic and static eOGMs, which are fed as inputs into our model. The model is trained on sequences of eOGMs, where a given sequence of $n$ frames consists of an input sequence of $k$ frames and a target sequence of $n-k$ frames. Our objective is to predict the future $n-k$ frames given the input sequence. Our approach is self-supervised as the target labels are eOGMs at future time instances. We outline these steps in further detail below. 

\subsection{eOGM Generation}
\label{approach:eOGM_gen}
We generate eOGMs following the process outlined in~\cite{Masha}. We discretize the local environment around the ego vehicle into grid cells that have the possibility of being either \textit{occupied} or \textit{free}, forming the frame of discernment: $\Omega = \{O, F\}$~\cite{dst}. The allowable hypotheses are those included in the power set of $\Omega$, with the exception of the empty set: $\{\{O\}, \{F\}, \{O, F\} \}$.
The empty set is not possible because a grid cell must be either occupied or free. We refer to $\{O, F\}$ as the \textit{occluded} set representing lack of information.
Each allowable hypothesis is associated with its corresponding Dempster--Shafer belief mass, which represents the degree of occupancy belief in that cell~\cite{dst}. The belief masses of all allowable hypotheses should sum to unity for each grid cell.

eOGMs are in $\mathbb{R}^{W \times H \times C}$, where $W$, $H$, and $C$ are the width, height, and number of channels. The channels consist of the DST belief masses for the occupied ($m(\{O\}) \in [0, 1]$) and free ($m(\{F\}) \in [0, 1]$) hypotheses. The mass for the occluded set can be computed using ${m(\{O,F\}) = 1 - m(\{O\}) - m(\{F\})}$. Prior to receiving sensor measurements, the eOGMs are initialized with $m(\{O,F\}) = 1$ ($m(\{O\}) = m(\{F\}) = 0$) for every cell to reflect lack of information. The belief mass, $m^c_t$, in cell $c$ at time step $t$ can be computed by fusing the previous belief mass, $m^c_{t-1}$, with the newly received sensor measurement, $m^c_{t,z}$~\cite{Nuss_DOGMa}, according to the Dempster--Shafer update rule~\cite{dst},
\begin{align}
\label{eq:update_rule}
m^c_t(A) & = (m^c_{t-1} \oplus m^c_{t,z})(A)\\
& \coloneqq \frac{\sum_{X \cap Y = A} m^c_{t-1}(X)m^c_{t,z}(Y)}
{1-\sum_{X \cap Y = \emptyset} m^c_{t-1}(X)m^c_{t,z}(Y)}, \nonumber \\
& \forall A,X,Y \in \{\{O\}, \{F\}, \{O,F\}\}, \nonumber
\end{align}
\noindent where $\oplus$ is the DST fusion operator. To account for information aging, we apply a discount factor to the previous belief masses. We construct OGMs for easier visual interpretability by converting the belief masses to estimated occupancy probabilities using the pignistic transformation as follows~\cite{pignistic},
\begin{align}
\label{eq:P_O}
p(O) = 0.5 \times m(\{O\}) + 0.5 \times (1-m(\{F\})).
\end{align}

\subsection{Dynamic Mask Generation}
\label{approach:dyn_mask}
We identify the moving objects in the environment to split the eOGMs into static and dynamic components. We assume that we have access to the detected objects and their tracking information, which is reasonable as AV perception systems traditionally include on-board real-time object detection and tracking capabilities.
Each detected object can then be classified as either moving or stationary by comparing their positions (bounding box centroids) between two consecutive frames. If the change in positions is more than a threshold, the object is determined to be moving. To account for the speed differences between object categories, we use different thresholds for pedestrians, cyclists, and vehicles. The sensor points corresponding to the bounding boxes classified as moving are then used to build the discretized dynamic masks, $M_{d} \in \mathbb{R}^{W \times H}$. The moving object cells are set to $1$, and $0$ otherwise. The dynamic eOGMs are obtained by multiplying the dynamic masks $M_{d}$ with the full eOGMs, whereas the static eOGMs are obtained by multiplying with $1-M_{d}$.

\subsection{Network Architecture}
The nature of the eOGM prediction task is spatiotemporal. Hence, our model uses a ConvLSTM~\cite{convlstm} architecture in the form of PredNet~\cite{PredNet}, following~\citet{Masha}, to learn both the temporal and spatial patterns in the eOGM data. However, our proposed double-prong model brings direct focus to the dynamic objects in the scenes, explicitly exploiting environment dynamics without using expensive particle filtering to estimate cell-wise velocities.

We construct a double-prong model where each prong consists of a reduced PredNet architecture~\cite{PredNet} from the original. The static prong takes as input the static eOGMs, which contain only the globally static portions of the environment, and learns to predict the future static eOGMs. We note that the static environment has local motion relative to the moving ego vehicle. Similarly, the dynamic prong takes the dynamic eOGMs as inputs and outputs dynamic eOGM predictions. The static and dynamic outputs are then fused to produce the full eOGM predictions using the Dempster--Shafer update rule~\cite{dst} (\cref{eq:update_rule}). Here, we combine the belief masses from the static, $m^c_{t,s}$, and dynamic, $m^c_{t,d}$, predictions for the same time instance: $m^c_{t}(A) = m^c_{t,s} \oplus m^c_{t,d}(A) \forall A \in \{\{O\}, \{F\}, \{O,F\}\}$. The OGMs are computed for analysis from the eOGMs using \cref{eq:P_O}.

\subsection{Training Loss}
The dynamic eOGMs are naturally highly sparse as there are often far fewer grid cells with moving objects than static objects and structures. Our training loss is thus a weighted combination between the loss from the dynamic predictions, $L_{d}$, and that from the full eOGM predictions, $L_{f}$,\linebreak
\begin{equation}
\label{eq:training_loss}
L = L_{d} + \alpha L_{f}.
\end{equation}
\noindent The full and dynamic eOGM losses per frame at a given time step are computed as the absolute errors (AE) in each cell $c$ between the target belief mass $m^{c}_{t}$ and the predicted belief mass $\hat{m}^{c}_{t}$ for the occupied ($\{O\}$) and free ($\{F\}$) channels,
\begin{align}
\label{eq:loss}
L_{t,f}(A) & = \frac{1}{W \times H} \sum_{c=1}^{W \times H} \mid m^{c}_{t}(A) - \hat{m}^{c}_{t}(A) \mid \\
L_{t,d}(A) & = \sum_{c=1}^{W \times H} \mid M_{d}^{c}m^{c}_{t}(A) - \hat{m}^{c}_{t, d}(A) \mid \\
& \forall A \in \{\{O\}, \{F\}\}, \nonumber
\end{align}
\noindent where $M_{d}$ is the dynamic mask. Here, $L_{t,f}(A)$ is the spatial mean of the AEs, whereas $L_{t,d}(A)$ is calculated as the spatial sum of the AEs. We find that this prevents the dynamic prong from simply predicting zero occupancy for every cell given the highly sparse dynamic eOGM inputs. The losses $L_{t,f}(A)$ and $L_{t,d}(A)$ are averaged over the channels and the time prediction horizon to compute $L_{f}$ and $L_{d}$, respectively.

\section{EXPERIMENTS}
\label{experiments}
\subsection{Data Generation}
We generate the eOGMs from LiDAR data in the Waymo Open Dataset~\cite{Sun_2020_CVPR}, which includes
a wide variety of scenes under different driving and weather conditions. The dataset contains $1,950$ driving segments of \SI{20}{\second} each collected at \SI{10}{\hertz}. The objects are tracked and labeled as vehicles, pedestrians, cyclists, or signs with associated $3$D bounding boxes and tracking IDs. To generate our training data, we subsample the Waymo dataset for frames that contain moving objects as, otherwise, the data is largely stationary, and we are interested in learning dynamic predictions.   

After filtering out the ground points from the LiDAR point clouds~\cite{Postica}, the eOGMs are generated as described in \cref{approach:eOGM_gen}. We construct eOGMs with a width and height of $128$ cells and a \SI{0.33}{\metre} resolution (approximately $\SI{42}{\metre} \times \SI{42}{\metre}$),
such that a reasonable FOV is covered, and there is a sufficient number of cells representing each vehicle. The ego vehicle is fixed at the center of every eOGM. Thus, the eOGMs represent the ego vehicle's frame of reference.

The moving components in the environment are identified as described in \cref{approach:dyn_mask}. We make use of the provided object labels and their tracking information to determine the changes in positions of the detected objects between frames. We empirically choose the speed thresholds: $\SI{1.4}{\m\per\s}$ for vehicles and $\SI{0.8}{\m\per\s}$ for pedestrians and cyclists. These values correspond to the approximate minimum speeds that we perceive the objects to be moving while also allowing for sensor reading errors. 
Since the data is sampled at $\SI{10}{\hertz}$, we classify the objects to be dynamic when the change in object positions between two consecutive frames is greater than $\SI{0.14}{\metre}$ for vehicles and $\SI{0.08}{\metre}$ for pedestrians and cyclists.
\begin{figure*}[t!]
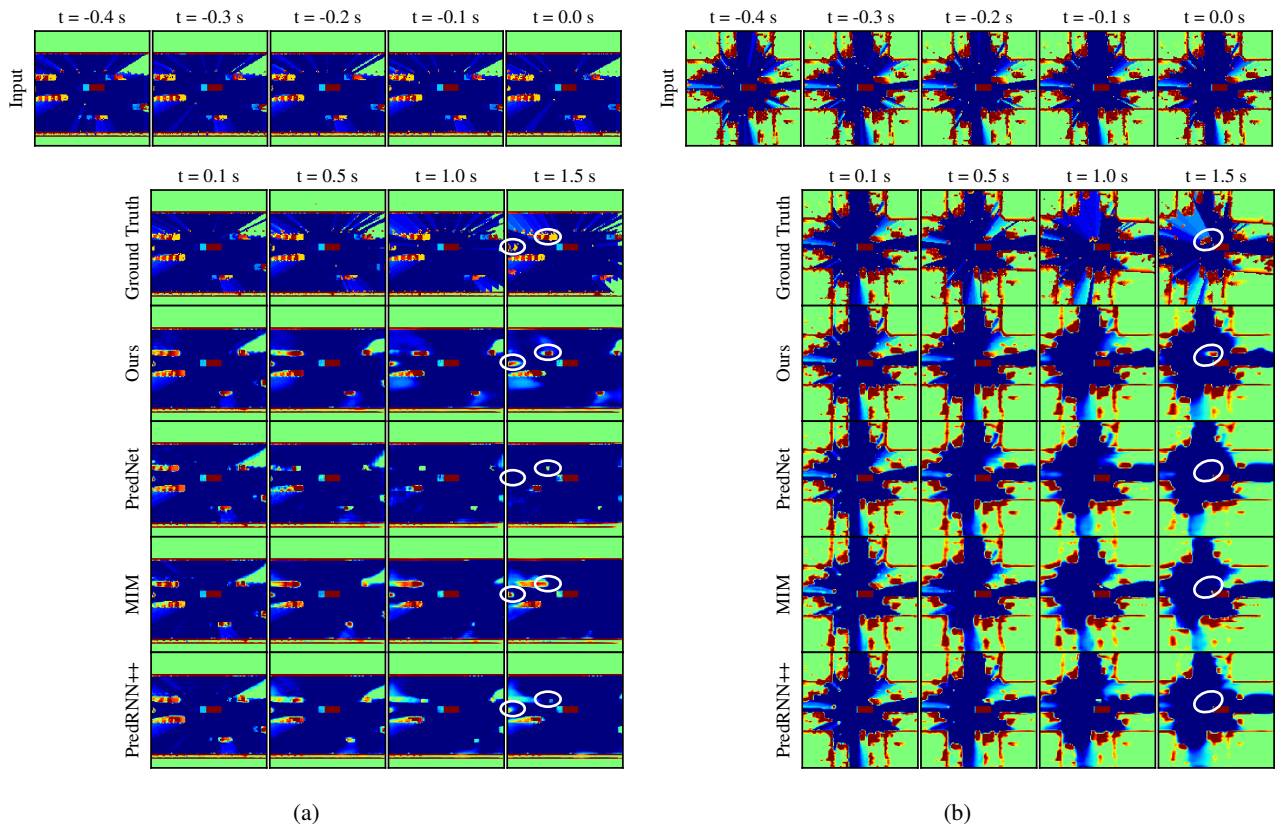

\centering
\begin{subfigure}[t]{0.48\textwidth}
\centering
\stackinset{l}{2.69in}{b}{2.81in}{\textcolor{white}{\solidcirc[0]{1pt}{0.16}{0.1}}}
{\stackinset{l}{2.69in}{b}{2.205in}{\textcolor{white}{\solidcirc[0]{1pt}{0.16}{0.1}}} 
{\stackinset{l}{2.69in}{b}{1.6in}{\textcolor{white}{\solidcirc[0]{1pt}{0.16}{0.1}}}
{\stackinset{l}{2.69in}{b}{0.99in}{\textcolor{white}{\solidcirc[0]{1pt}{0.16}{0.1}}}
{\stackinset{l}{2.69in}{b}{0.39in}{\textcolor{white}{\solidcirc[0]{1pt}{0.16}{0.1}}}
{\stackinset{l}{2.87in}{b}{2.862in}{\textcolor{white}{\solidcirc[0]{1pt}{0.17}{0.1}}}
{\stackinset{l}{2.87in}{b}{2.257in}{\textcolor{white}{\solidcirc[0]{1pt}{0.17}{0.1}}} 
{\stackinset{l}{2.87in}{b}{1.65in}{\textcolor{white}{\solidcirc[0]{1pt}{0.17}{0.1}}}
{\stackinset{l}{2.87in}{b}{1.045in}{\textcolor{white}{\solidcirc[0]{1pt}{0.17}{0.1}}}
{\stackinset{l}{2.87in}{b}{0.44in}{\textcolor{white}{\solidcirc[0]{1pt}{0.17}{0.1}}}
{\input{Figures/Qualitative/qualitative_straight}}}}}}}}}}}
\vspace{-0.5cm}
\caption{}
\label{fig:quali_straight}
\end{subfigure}
\begin{subfigure}[t]{0.48\textwidth}
\centering
\stackinset{l}{2.895in}{b}{2.81in}{\textcolor{white}{\solidcirc[20]{1pt}{0.192}{0.13}}}
{\stackinset{l}{2.895in}{b}{2.205in}{\textcolor{white}{\solidcirc[20]{1pt}{0.192}{0.13}}} 
{\stackinset{l}{2.895in}{b}{1.6in}{\textcolor{white}{\solidcirc[20]{1pt}{0.192}{0.13}}}
{\stackinset{l}{2.895in}{b}{0.99in}{\textcolor{white}{\solidcirc[20]{1pt}{0.192}{0.13}}}
{\stackinset{l}{2.895in}{b}{0.39in}{\textcolor{white}{\solidcirc[20]{1pt}{0.192}{0.13}}}
{\input{Figures/Qualitative/qualitative_turning}}}}}}
\vspace{-0.5cm}
\caption{}
\label{fig:quali_turning}
\end{subfigure}
\caption{\small Example OGM predictions (red: occupied, green: occluded, blue: free) at of \SI{0.1}{\second}, \SI{0.5}{\second}, \SI{1.0}{\second}, and \SI{1.5}{\second} ahead. \cref{fig:quali_straight} shows an example scene with multiple vehicles moving straight at high speeds. \cref{fig:quali_turning} shows an example scene of multiple vehicles moving through an intersection. In both scenes, our model retains dynamic objects in the prediction for longer prediction horizons than the baselines.}
\vspace{-0.55cm}
\label{fig:qualitative}
\end{figure*}
\subsection{Experimental Details and Baselines}
The inputs to our double-prong model are the static and dynamic eOGMs and the outputs are the complete eOGM predictions. 
We arrange the eOGMs into sequences of $20$ consecutive frames each, representing $\SI{2}{\s}$ of driving data.
The model is trained to predict the future $15$ eOGM frames ($\SI{1.5}{\s}$) based on an input of the past $5$ eOGM frames ($\SI{0.5}{\s}$) in the same sequence. 
We have $3061$, $780$, and $1071$ sequences for training, validation, and testing, respectively. Before training the model to predict the next $15$ frames from the past $5$ frames, we first train the model to predict the eOGM at the next time step ($\SI{0.1}{\s}$) given the current eOGM, as suggested by \citet{PredNet}. The model is then fine-tuned to recursively predict the next $15$ eOGM frames by initializing the weight parameters from the previous training. The static and dynamic prongs consist of $3$ and $2$ PredNet layers, respectively. We use $2$-dilated convolution in the second layer of the dynamic prong. The network is trained using the Adam optimizer~\cite{adam} with a starting learning rate of $0.00001$ for a total of $60$ epochs with $2000$ samples per epoch for each training stage. We found that removing the denominator in the Dempster--Shafer update rule (\cref{eq:update_rule}) when fusing the static and dynamic predictions results in better training stability. We adjust for this normalization post-training when evaluating our model. We set $\alpha = 10$ in the loss (\cref{eq:training_loss}) based on validation set performance.

We baseline our proposed double-prong model results against three video frame prediction architectures, namely PredNet~\cite{PredNet}, MIM~\cite{mim}, and PredRNN++~\cite{predrnn++}. The number of layers in the baseline models is reduced from the original architecture to $3$ so that the number of parameters is of the same order of magnitude as that of our model for fair comparison. 

\section{RESULTS}
\label{results}
This section presents the findings from our proposed model and considered baselines for environment prediction. We perform our analysis on OGMs, obtained using \cref{eq:P_O}, rather than eOGMs for easier interpretability.\footnote{Our implementation is available at: \url{https://github.com/sisl/Double-Prong-Occupancy}.}
\subsection{Qualitative Results}
\cref{fig:qualitative} illustrates example predictions for two driving scenes using our proposed double-prong model as compared to the PredNet~\cite{PredNet}, MIM~\cite{mim}, and PredRNN++~\cite{predrnn++} baselines. The top row depicts the $5$~OGMs that the models receive as inputs. The ground truth labels show target OGMs at selected prediction times.
The ego vehicle is fixed at the center and is traveling to the right in the OGMs. Since the OGMs are egocentric, the motion of other objects in the environment is relative to the ego vehicle. Occupied, occluded, and free cells are depicted in red, green, and blue, respectively. 

\cref{fig:quali_straight} shows a high-speed scene with vehicles moving faster than the ego vehicle. There are six vehicles present in the scene, excluding the ego, that are heading to the right in the OGMs. Our model is able to retain all the vehicles in the predictions and accurately capture their motions. 

We first consider the three vehicles behind the ego vehicle (left side in the OGMs). Two of these vehicles are shown with long trails, representing uncertain information and indicating that they are moving at relatively high speeds. The third vehicle only starts to appear in the frame in the input OGMs. Our model retains these vehicles for the \SI{1.5}{\second} prediction horizon, and accurately predicts their motions. It is able to extrapolate the shape and speed of the partially out-of-range vehicle (circled in the OGMs). We hypothesize that similar examples of partially out-of-range vehicles appeared in the training data, allowing the network to learn this behavior. Although the MIM network, which is more than three times larger than our model, also retains these vehicles, it does not capture the motion of the top vehicle (circled in the OGMs) as accurately. Both PredNet and PredRNN++ fail to retain the top vehicle in their predictions.

We now consider the remaining three vehicles in front of the ego vehicle (right side in the OGMs). The middle vehicle moves beyond the borders of the OGMs (in ground truth frames at $t =$ \SI{1.0}{\second}, \SI{1.5}{\second}). Our model accurately predicts the motion of these vehicles, including the vehicle that moves out of the frame, whereas they vanish in the baseline predictions.

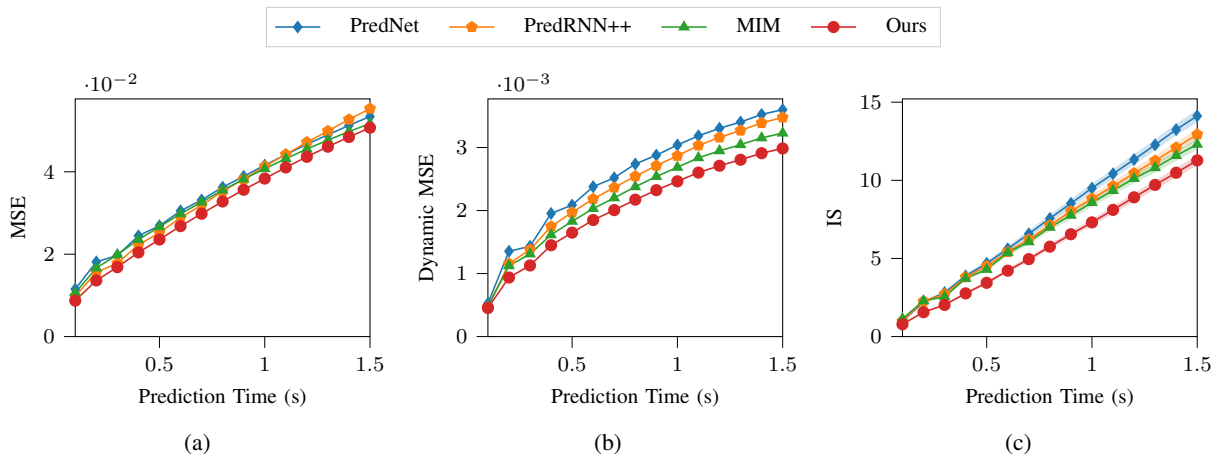
\begin{figure*}[t!]
    \centering
    \begin{subfigure}[t]{0.8\textwidth}
        \centering
        \begin{tikzpicture}
\definecolor{color0}{rgb}{0.12156862745098,0.466666666666667,0.705882352941177}
\definecolor{color1}{rgb}{1,0.498039215686275,0.0549019607843137}
\definecolor{color2}{rgb}{0.172549019607843,0.627450980392157,0.172549019607843}
\definecolor{color3}{rgb}{0.83921568627451,0.152941176470588,0.156862745098039}

\begin{customlegend}[legend columns=4,legend style={align=left,draw=white!80!black,column sep=2ex},
        legend entries={\footnotesize PredNet,
                        \footnotesize PredRNN++,
                        \footnotesize MIM,
                        \footnotesize Ours}]
        \addlegendimage{semithick, color0, mark=diamond*, mark size=2, mark options={solid}}
        \addlegendimage{semithick, color1, mark=pentagon*, mark size=2, mark options={solid}}
        \addlegendimage{semithick, color2, mark=triangle*, mark size=2, mark options={solid}}
        \addlegendimage{semithick, color3, mark=*, mark size=2, mark options={solid}}
        \end{customlegend}
\end{tikzpicture}\vspace{0.35cm}
        \label{fig:legend}
    \end{subfigure}
    \begin{subfigure}[t]{0.3\textwidth}
        \centering
\begin{tikzpicture}

\definecolor{color0}{rgb}{0.12156862745098,0.466666666666667,0.705882352941177}
\definecolor{color1}{rgb}{1,0.498039215686275,0.0549019607843137}
\definecolor{color2}{rgb}{0.172549019607843,0.627450980392157,0.172549019607843}
\definecolor{color3}{rgb}{0.83921568627451,0.152941176470588,0.156862745098039}

\begin{axis}[
width = 5.5cm,
tick align=outside,
tick pos=left,
x grid style={white!69.0196078431373!black},
xlabel={\footnotesize Prediction Time (s)},
xmin=0.1, xmax=1.5,
xtick style={color=black},
y grid style={white!69.0196078431373!black},
ylabel={\footnotesize MSE},
ymin=0, ymax=0.0576937549747527,
ytick style={color=black}
]
\path [fill=color0, fill opacity=0.2]
(axis cs:0.1,0.0115230260416865)
--(axis cs:0.1,0.0114949373528361)
--(axis cs:0.2,0.0181152708828449)
--(axis cs:0.3,0.0196706634014845)
--(axis cs:0.4,0.0244391262531281)
--(axis cs:0.5,0.0268847979605198)
--(axis cs:0.6,0.0304882060736418)
--(axis cs:0.7,0.0331321321427822)
--(axis cs:0.8,0.0362310074269772)
--(axis cs:0.9,0.0388573557138443)
--(axis cs:1,0.0415804609656334)
--(axis cs:1.1,0.0441251881420612)
--(axis cs:1.2,0.0466493964195251)
--(axis cs:1.3,0.0489403009414673)
--(axis cs:1.4,0.0512288101017475)
--(axis cs:1.5,0.0533789247274399)
--(axis cs:1.5,0.0534522980451584)
--(axis cs:1.5,0.0534522980451584)
--(axis cs:1.4,0.0513005815446377)
--(axis cs:1.3,0.0490102469921112)
--(axis cs:1.2,0.0467175170779228)
--(axis cs:1.1,0.0441911928355694)
--(axis cs:1,0.0416443049907684)
--(axis cs:0.9,0.0389187857508659)
--(axis cs:0.8,0.0362900383770466)
--(axis cs:0.7,0.0331881679594517)
--(axis cs:0.6,0.0305415447801352)
--(axis cs:0.5,0.0269343070685863)
--(axis cs:0.4,0.0244858674705029)
--(axis cs:0.3,0.0197114031761885)
--(axis cs:0.2,0.0181538127362728)
--(axis cs:0.1,0.0115230260416865)
--cycle;

\path [fill=color1, fill opacity=0.2]
(axis cs:0.1,0.00980930123478174)
--(axis cs:0.1,0.00978267472237349)
--(axis cs:0.2,0.0155892707407475)
--(axis cs:0.3,0.0177780874073505)
--(axis cs:0.4,0.0222625713795424)
--(axis cs:0.5,0.0251630898565054)
--(axis cs:0.6,0.0288959313184023)
--(axis cs:0.7,0.0318387448787689)
--(axis cs:0.8,0.0352144949138165)
--(axis cs:0.9,0.038164459168911)
--(axis cs:1,0.0412772111594677)
--(axis cs:1.1,0.0441783592104912)
--(axis cs:1.2,0.0471462234854698)
--(axis cs:1.3,0.0498691201210022)
--(axis cs:1.4,0.0526591539382935)
--(axis cs:1.5,0.0552848540246487)
--(axis cs:1.5,0.0553612746298313)
--(axis cs:1.5,0.0553612746298313)
--(axis cs:1.4,0.0527335181832314)
--(axis cs:1.3,0.0499411970376968)
--(axis cs:1.2,0.0472160875797272)
--(axis cs:1.1,0.0442457050085068)
--(axis cs:1,0.0413420312106609)
--(axis cs:0.9,0.038226418197155)
--(axis cs:0.8,0.0352736748754978)
--(axis cs:0.7,0.0318945795297623)
--(axis cs:0.6,0.0289487261325121)
--(axis cs:0.5,0.0252117868512869)
--(axis cs:0.4,0.0223078448325396)
--(axis cs:0.3,0.0178174562752247)
--(axis cs:0.2,0.0156253706663847)
--(axis cs:0.1,0.00980930123478174)
--cycle;

\path [fill=color2, fill opacity=0.2]
(axis cs:0.1,0.0106331668794155)
--(axis cs:0.1,0.010606300085783)
--(axis cs:0.2,0.0166284833103418)
--(axis cs:0.3,0.0198370330035686)
--(axis cs:0.4,0.0235307365655899)
--(axis cs:0.5,0.0266188997775316)
--(axis cs:0.6,0.029783496633172)
--(axis cs:0.7,0.0326156727969646)
--(axis cs:0.8,0.0355178564786911)
--(axis cs:0.9,0.0381776094436646)
--(axis cs:1,0.0407124385237694)
--(axis cs:1.1,0.0431488081812859)
--(axis cs:1.2,0.0454875826835632)
--(axis cs:1.3,0.0476593561470509)
--(axis cs:1.4,0.0497311502695084)
--(axis cs:1.5,0.0516629181802273)
--(axis cs:1.5,0.0517335496842861)
--(axis cs:1.5,0.0517335496842861)
--(axis cs:1.4,0.049800418317318)
--(axis cs:1.3,0.0477270893752575)
--(axis cs:1.2,0.0455536916851997)
--(axis cs:1.1,0.0432130917906761)
--(axis cs:1,0.040774755179882)
--(axis cs:0.9,0.0382377654314041)
--(axis cs:0.8,0.0355756506323814)
--(axis cs:0.7,0.0326707549393177)
--(axis cs:0.6,0.0298357885330915)
--(axis cs:0.5,0.0266679208725691)
--(axis cs:0.4,0.0235761851072311)
--(axis cs:0.3,0.0198779925704002)
--(axis cs:0.2,0.0166649948805571)
--(axis cs:0.1,0.0106331668794155)
--cycle;

\path [fill=color3, fill opacity=0.2]
(axis cs:0.1,0.00873587839305401)
--(axis cs:0.1,0.0087116677314043)
--(axis cs:0.2,0.0136114815250039)
--(axis cs:0.3,0.0168339628726244)
--(axis cs:0.4,0.0204042606055737)
--(axis cs:0.5,0.0235336758196354)
--(axis cs:0.6,0.0267926622182131)
--(axis cs:0.7,0.0297941286116838)
--(axis cs:0.8,0.0327679365873337)
--(axis cs:0.9,0.0356080271303654)
--(axis cs:1,0.0383266322314739)
--(axis cs:1.1,0.0410220101475716)
--(axis cs:1.2,0.0436085425317287)
--(axis cs:1.3,0.0460593737661839)
--(axis cs:1.4,0.0484111346304417)
--(axis cs:1.5,0.0506624206900597)
--(axis cs:1.5,0.0507335588335991)
--(axis cs:1.5,0.0507335588335991)
--(axis cs:1.4,0.0484804920852184)
--(axis cs:1.3,0.0461268462240696)
--(axis cs:1.2,0.0436739958822727)
--(axis cs:1.1,0.0410852432250977)
--(axis cs:1,0.038387481123209)
--(axis cs:0.9,0.0356663502752781)
--(axis cs:0.8,0.0328235477209091)
--(axis cs:0.7,0.029846778139472)
--(axis cs:0.6,0.0268421266227961)
--(axis cs:0.5,0.0235795564949512)
--(axis cs:0.4,0.0204463824629784)
--(axis cs:0.3,0.0168713796883821)
--(axis cs:0.2,0.0136440107598901)
--(axis cs:0.1,0.00873587839305401)
--cycle;

\addplot [semithick, color0, mark=diamond*, mark size=2, mark options={solid}]
table {%
0.1 0.0115089816972613
0.2 0.0181345418095589
0.3 0.0196910332888365
0.4 0.0244624968618155
0.5 0.0269095525145531
0.6 0.0305148754268885
0.7 0.0331601500511169
0.8 0.0362605229020119
0.9 0.0388880707323551
1 0.0416123829782009
1.1 0.0441581904888153
1.2 0.046683456748724
1.3 0.0489752739667892
1.4 0.0512646958231926
1.5 0.0534156113862991
};

\addplot [semithick, color1, mark=pentagon*, mark size=2, mark options={solid}]
table {%
0.1 0.00979598797857761
0.2 0.0156073207035661
0.3 0.0177977718412876
0.4 0.022285208106041
0.5 0.0251874383538961
0.6 0.0289223287254572
0.7 0.0318666622042656
0.8 0.0352440848946571
0.9 0.038195438683033
1 0.0413096211850643
1.1 0.044212032109499
1.2 0.0471811555325985
1.3 0.0499051585793495
1.4 0.0526963360607624
1.5 0.05532306432724
};

\addplot [semithick, color2, mark=triangle*, mark size=2, mark options={solid}]
table {%
0.1 0.0106197334825993
0.2 0.0166467390954494
0.3 0.0198575127869844
0.4 0.0235534608364105
0.5 0.0266434103250504
0.6 0.0298096425831318
0.7 0.0326432138681412
0.8 0.0355467535555363
0.9 0.0382076874375343
1 0.0407435968518257
1.1 0.043180949985981
1.2 0.0455206371843815
1.3 0.0476932227611542
1.4 0.0497657842934132
1.5 0.0516982339322567
};

\addplot [semithick, color3, mark=*, mark size=2, mark options={solid}]
table {%
0.1 0.00872377306222916
0.2 0.013627746142447
0.3 0.0168526712805033
0.4 0.020425321534276
0.5 0.0235566161572933
0.6 0.0268173944205046
0.7 0.0298204533755779
0.8 0.0327957421541214
0.9 0.0356371887028217
1 0.0383570566773415
1.1 0.0410536266863346
1.2 0.0436412692070007
1.3 0.0460931099951267
1.4 0.04844581335783
1.5 0.0506979897618294
};

\end{axis}

\end{tikzpicture}\\
        \caption{}
        \label{fig:mse_full}
    \end{subfigure} 
    \begin{subfigure}[t]{0.3\textwidth}
        \centering
\begin{tikzpicture}

\definecolor{color0}{rgb}{0.12156862745098,0.466666666666667,0.705882352941177}
\definecolor{color1}{rgb}{1,0.498039215686275,0.0549019607843137}
\definecolor{color2}{rgb}{0.172549019607843,0.627450980392157,0.172549019607843}
\definecolor{color3}{rgb}{0.83921568627451,0.152941176470588,0.156862745098039}

\begin{axis}[
width = 5.5cm,
tick align=outside,
tick pos=left,
x grid style={white!69.0196078431373!black},
xlabel={\footnotesize Prediction Time (s)},
xmin=0.1, xmax=1.5,
xtick style={color=black},
y grid style={white!69.0196078431373!black},
ylabel={\footnotesize Dynamic MSE},
ymin=0, ymax=0.00377132888970664,
ytick style={color=black}
]
\path [fill=color0, fill opacity=0.2]
(axis cs:0.1,0.00053002720233053)
--(axis cs:0.1,0.000523101887665689)
--(axis cs:0.2,0.00134680175688118)
--(axis cs:0.3,0.00142457312904298)
--(axis cs:0.4,0.00194511644076556)
--(axis cs:0.5,0.00207727123051882)
--(axis cs:0.6,0.00236887251958251)
--(axis cs:0.7,0.00250927545130253)
--(axis cs:0.8,0.00272787897847593)
--(axis cs:0.9,0.0028677424415946)
--(axis cs:1,0.00302819558419287)
--(axis cs:1.1,0.00317365187220275)
--(axis cs:1.2,0.00329321390017867)
--(axis cs:1.3,0.00339196040295064)
--(axis cs:1.4,0.00351203116588295)
--(axis cs:1.5,0.00358793861232698)
--(axis cs:1.5,0.0036131648812443)
--(axis cs:1.5,0.0036131648812443)
--(axis cs:1.4,0.0035369258839637)
--(axis cs:1.3,0.00341632938943803)
--(axis cs:1.2,0.00331715308129787)
--(axis cs:1.1,0.00319702574051917)
--(axis cs:1,0.00305093475617468)
--(axis cs:0.9,0.00288974121212959)
--(axis cs:0.8,0.00274925003759563)
--(axis cs:0.7,0.00252957595512271)
--(axis cs:0.6,0.00238847639411688)
--(axis cs:0.5,0.00209530536085367)
--(axis cs:0.4,0.00196250923909247)
--(axis cs:0.3,0.00143867102451622)
--(axis cs:0.2,0.00136052479501814)
--(axis cs:0.1,0.00053002720233053)
--cycle;

\path [fill=color1, fill opacity=0.2]
(axis cs:0.1,0.000475273962365463)
--(axis cs:0.1,0.000469442718895152)
--(axis cs:0.2,0.00115205685142428)
--(axis cs:0.3,0.00137309404090047)
--(axis cs:0.4,0.00173959718085825)
--(axis cs:0.5,0.00196230388246477)
--(axis cs:0.6,0.00217590807005763)
--(axis cs:0.7,0.00235577928833663)
--(axis cs:0.8,0.00253469287417829)
--(axis cs:0.9,0.00270129204727709)
--(axis cs:1,0.0028581558726728)
--(axis cs:1.1,0.00302157201804221)
--(axis cs:1.2,0.00314886821433902)
--(axis cs:1.3,0.00325826928019524)
--(axis cs:1.4,0.00337789626792073)
--(axis cs:1.5,0.00346409156918526)
--(axis cs:1.5,0.00348852667957544)
--(axis cs:1.5,0.00348852667957544)
--(axis cs:1.4,0.00340191507712007)
--(axis cs:1.3,0.00328174233436584)
--(axis cs:1.2,0.00317178433761001)
--(axis cs:1.1,0.00304385996423662)
--(axis cs:1,0.00287965079769492)
--(axis cs:0.9,0.00272202887572348)
--(axis cs:0.8,0.00255459663458169)
--(axis cs:0.7,0.00237478618510067)
--(axis cs:0.6,0.00219392543658614)
--(axis cs:0.5,0.0019791426602751)
--(axis cs:0.4,0.00175513094291091)
--(axis cs:0.3,0.00138624897226691)
--(axis cs:0.2,0.00116373121272773)
--(axis cs:0.1,0.000475273962365463)
--cycle;

\path [fill=color2, fill opacity=0.2]
(axis cs:0.1,0.00048492118366994)
--(axis cs:0.1,0.000478154077427462)
--(axis cs:0.2,0.00111675902735442)
--(axis cs:0.3,0.0013036368181929)
--(axis cs:0.4,0.00160695682279766)
--(axis cs:0.5,0.00182029453571886)
--(axis cs:0.6,0.00202225893735886)
--(axis cs:0.7,0.00219114939682186)
--(axis cs:0.8,0.00236699637025595)
--(axis cs:0.9,0.00252965511754155)
--(axis cs:1,0.00267470069229603)
--(axis cs:1.1,0.00282629788853228)
--(axis cs:1.2,0.00293601700104773)
--(axis cs:1.3,0.003033327171579)
--(axis cs:1.4,0.00314064580015838)
--(axis cs:1.5,0.00321642169728875)
--(axis cs:1.5,0.00324026495218277)
--(axis cs:1.5,0.00324026495218277)
--(axis cs:1.4,0.00316413375549018)
--(axis cs:1.3,0.00305634248070419)
--(axis cs:1.2,0.00295857456512749)
--(axis cs:1.1,0.00284834648482502)
--(axis cs:1,0.00269602751359344)
--(axis cs:0.9,0.00255027320235968)
--(axis cs:0.8,0.00238679256290197)
--(axis cs:0.7,0.00221007945947349)
--(axis cs:0.6,0.0020402786321938)
--(axis cs:0.5,0.00183721689973027)
--(axis cs:0.4,0.00162262655794621)
--(axis cs:0.3,0.00131741969380528)
--(axis cs:0.2,0.00112910533789545)
--(axis cs:0.1,0.00048492118366994)
--cycle;

\path [fill=color3, fill opacity=0.2]
(axis cs:0.1,0.000456094305263832)
--(axis cs:0.1,0.000449884711997584)
--(axis cs:0.2,0.000931741204112768)
--(axis cs:0.3,0.0011224530171603)
--(axis cs:0.4,0.00144232425373048)
--(axis cs:0.5,0.00164140842389315)
--(axis cs:0.6,0.0018408497562632)
--(axis cs:0.7,0.00199808506295085)
--(axis cs:0.8,0.0021632038988173)
--(axis cs:0.9,0.00231356685981154)
--(axis cs:1,0.00245389924384654)
--(axis cs:1.1,0.00259184092283249)
--(axis cs:1.2,0.00269995164126158)
--(axis cs:1.3,0.00279516959562898)
--(axis cs:1.4,0.0028984728269279)
--(axis cs:1.5,0.00297358143143356)
--(axis cs:1.5,0.0029961506370455)
--(axis cs:1.5,0.0029961506370455)
--(axis cs:1.4,0.00292067881673574)
--(axis cs:1.3,0.00281690526753664)
--(axis cs:1.2,0.00272120768204331)
--(axis cs:1.1,0.00261255772784352)
--(axis cs:1,0.00247395015321672)
--(axis cs:0.9,0.00233289739117026)
--(axis cs:0.8,0.00218172464519739)
--(axis cs:0.7,0.00201570475474)
--(axis cs:0.6,0.00185755791608244)
--(axis cs:0.5,0.00165691704023629)
--(axis cs:0.4,0.00145659444388002)
--(axis cs:0.3,0.00113440724089742)
--(axis cs:0.2,0.000942283775657415)
--(axis cs:0.1,0.000456094305263832)
--cycle;

\addplot [semithick, color0, mark=diamond*, mark size=2, mark options={solid}]
table {%
0.1 0.000526564544998109
0.2 0.00135366327594966
0.3 0.0014316220767796
0.4 0.00195381278172135
0.5 0.00208628829568624
0.6 0.00237867445684969
0.7 0.00251942570321262
0.8 0.00273856450803578
0.9 0.0028787418268621
1 0.00303956517018378
1.1 0.00318533880636096
1.2 0.00330518349073827
1.3 0.00340414489619434
1.4 0.00352447852492332
1.5 0.00360055174678564
};

\addplot [semithick, color1, mark=pentagon*, mark size=2, mark options={solid}]
table {%
0.1 0.000472358340630308
0.2 0.001157894032076
0.3 0.00137967150658369
0.4 0.00174736406188458
0.5 0.00197072327136993
0.6 0.00218491675332189
0.7 0.00236528273671865
0.8 0.00254464475437999
0.9 0.00271166046150029
1 0.00286890333518386
1.1 0.00303271599113941
1.2 0.00316032627597451
1.3 0.00327000580728054
1.4 0.0033899056725204
1.5 0.00347630912438035
};

\addplot [semithick, color2, mark=triangle*, mark size=2, mark options={solid}]
table {%
0.1 0.000481537630548701
0.2 0.00112293218262494
0.3 0.00131052825599909
0.4 0.00161479169037193
0.5 0.00182875571772456
0.6 0.00203126878477633
0.7 0.00220061442814767
0.8 0.00237689446657896
0.9 0.00253996415995061
1 0.00268536410294473
1.1 0.00283732218667865
1.2 0.00294729578308761
1.3 0.0030448348261416
1.4 0.00315238977782428
1.5 0.00322834332473576
};

\addplot [semithick, color3, mark=*, mark size=2, mark options={solid}]
table {%
0.1 0.000452989508630708
0.2 0.000937012489885092
0.3 0.00112843012902886
0.4 0.00144945934880525
0.5 0.00164916273206472
0.6 0.00184920383617282
0.7 0.00200689490884542
0.8 0.00217246427200735
0.9 0.0023232321254909
1 0.00246392469853163
1.1 0.00260219932533801
1.2 0.00271057966165245
1.3 0.00280603743158281
1.4 0.00290957582183182
1.5 0.00298486603423953
};

\end{axis}

\end{tikzpicture}\\
        \caption{}
        \label{fig:mse_dyn}
    \end{subfigure}
    \begin{subfigure}[t]{0.3\textwidth}
        \centering
\begin{tikzpicture}

\definecolor{color0}{rgb}{0.12156862745098,0.466666666666667,0.705882352941177}
\definecolor{color1}{rgb}{1,0.498039215686275,0.0549019607843137}
\definecolor{color2}{rgb}{0.172549019607843,0.627450980392157,0.172549019607843}
\definecolor{color3}{rgb}{0.83921568627451,0.152941176470588,0.156862745098039}

\begin{axis}[
width=5.5cm,
tick align=outside,
tick pos=left,
x grid style={white!69.0196078431373!black},
xlabel={\footnotesize Prediction Time (s)},
xmin=0.1, xmax=1.5,
xtick style={color=black},
y grid style={white!69.0196078431373!black},
ylabel={\footnotesize IS},
ymin=0, ymax=15.2096629642245,
ytick style={color=black}
]
\path [fill=color0, fill opacity=0.2]
(axis cs:0.1,1.01641293992651)
--(axis cs:0.1,0.950790523757227)
--(axis cs:0.2,2.12129134837162)
--(axis cs:0.3,2.71215598383501)
--(axis cs:0.4,3.74641900018425)
--(axis cs:0.5,4.52356396480791)
--(axis cs:0.6,5.42683100265639)
--(axis cs:0.7,6.35728966194184)
--(axis cs:0.8,7.31398496458472)
--(axis cs:0.9,8.25292356495107)
--(axis cs:1,9.20048789788866)
--(axis cs:1.1,10.0999520351681)
--(axis cs:1.2,10.9877495650045)
--(axis cs:1.3,11.9095803711313)
--(axis cs:1.4,12.8605132438478)
--(axis cs:1.5,13.7184431785018)
--(axis cs:1.5,14.5217339604843)
--(axis cs:1.5,14.5217339604843)
--(axis cs:1.4,13.626977136436)
--(axis cs:1.3,12.6324726886687)
--(axis cs:1.2,11.6673934517937)
--(axis cs:1.1,10.7384723639243)
--(axis cs:1,9.79920371779193)
--(axis cs:0.9,8.79842295846552)
--(axis cs:0.8,7.80720357646648)
--(axis cs:0.7,6.79707763728408)
--(axis cs:0.6,5.80673950758039)
--(axis cs:0.5,4.84852591099116)
--(axis cs:0.4,4.02353202064037)
--(axis cs:0.3,2.9155473691888)
--(axis cs:0.2,2.28159279845082)
--(axis cs:0.1,1.01641293992651)
--cycle;

\path [fill=color1, fill opacity=0.2]
(axis cs:0.1,1.02399659551139)
--(axis cs:0.1,0.956143369152755)
--(axis cs:0.2,2.1585377203086)
--(axis cs:0.3,2.60110079050564)
--(axis cs:0.4,3.64921608734014)
--(axis cs:0.5,4.34028835830308)
--(axis cs:0.6,5.22194850288402)
--(axis cs:0.7,5.99766462486507)
--(axis cs:0.8,6.8871846317312)
--(axis cs:0.9,7.75475641621734)
--(axis cs:1,8.53996871368427)
--(axis cs:1.1,9.32881086665427)
--(axis cs:1.2,10.1358675051224)
--(axis cs:1.3,10.9229656256432)
--(axis cs:1.4,11.7469756757792)
--(axis cs:1.5,12.5705608734596)
--(axis cs:1.5,13.2969177598578)
--(axis cs:1.5,13.2969177598578)
--(axis cs:1.4,12.4450986726319)
--(axis cs:1.3,11.587640316236)
--(axis cs:1.2,10.7777017551676)
--(axis cs:1.1,9.92820145281143)
--(axis cs:1,9.10927234269984)
--(axis cs:0.9,8.28743567240122)
--(axis cs:0.8,7.36049262676847)
--(axis cs:0.7,6.4200575564887)
--(axis cs:0.6,5.60786160324038)
--(axis cs:0.5,4.67797476574217)
--(axis cs:0.4,3.96080521118518)
--(axis cs:0.3,2.81634587132629)
--(axis cs:0.2,2.36127019379106)
--(axis cs:0.1,1.02399659551139)
--cycle;

\path [fill=color2, fill opacity=0.2]
(axis cs:0.1,1.17792043262392)
--(axis cs:0.1,1.10908662059262)
--(axis cs:0.2,2.21097980522238)
--(axis cs:0.3,2.4645265275422)
--(axis cs:0.4,3.6045861381355)
--(axis cs:0.5,4.16024978086174)
--(axis cs:0.6,5.19216654188757)
--(axis cs:0.7,5.86399697419025)
--(axis cs:0.8,6.76605774882774)
--(axis cs:0.9,7.51294102822295)
--(axis cs:1,8.30826381943292)
--(axis cs:1.1,9.05598515122128)
--(axis cs:1.2,9.80849955579342)
--(axis cs:1.3,10.4866252495378)
--(axis cs:1.4,11.2432373848843)
--(axis cs:1.5,11.9484959097639)
--(axis cs:1.5,12.6559279809556)
--(axis cs:1.5,12.6559279809556)
--(axis cs:1.4,11.9176409861664)
--(axis cs:1.3,11.1193847586794)
--(axis cs:1.2,10.4175676941021)
--(axis cs:1.1,9.62865083411668)
--(axis cs:1,8.84331511944751)
--(axis cs:0.9,8.00405970032351)
--(axis cs:0.8,7.21706303947246)
--(axis cs:0.7,6.25982121494433)
--(axis cs:0.6,5.54426985273069)
--(axis cs:0.5,4.43388581142364)
--(axis cs:0.4,3.84582426445359)
--(axis cs:0.3,2.62419527812309)
--(axis cs:0.2,2.35947216412375)
--(axis cs:0.1,1.17792043262392)
--cycle;

\path [fill=color3, fill opacity=0.2]
(axis cs:0.1,0.80562604455758)
--(axis cs:0.1,0.763153885681791)
--(axis cs:0.2,1.5138378849815)
--(axis cs:0.3,1.96230330736174)
--(axis cs:0.4,2.68111715886555)
--(axis cs:0.5,3.33484743236872)
--(axis cs:0.6,4.087171084079)
--(axis cs:0.7,4.8046480925321)
--(axis cs:0.8,5.57807456394567)
--(axis cs:0.9,6.3500726663618)
--(axis cs:1,7.08794926630058)
--(axis cs:1.1,7.86394456045051)
--(axis cs:1.2,8.6463705576689)
--(axis cs:1.3,9.42104775139447)
--(axis cs:1.4,10.1801743414475)
--(axis cs:1.5,10.9447856369191)
--(axis cs:1.5,11.6018207697797)
--(axis cs:1.5,11.6018207697797)
--(axis cs:1.4,10.7886509330637)
--(axis cs:1.3,9.99199154940565)
--(axis cs:1.2,9.17697529570762)
--(axis cs:1.1,8.35110390399104)
--(axis cs:1,7.52693373801035)
--(axis cs:0.9,6.74826881170732)
--(axis cs:0.8,5.93092916318737)
--(axis cs:0.7,5.10883435961388)
--(axis cs:0.6,4.34710622051481)
--(axis cs:0.5,3.54565041127754)
--(axis cs:0.4,2.84788102845591)
--(axis cs:0.3,2.08530213223367)
--(axis cs:0.2,1.60948496488107)
--(axis cs:0.1,0.80562604455758)
--cycle;

\addplot [semithick, color0, mark=diamond*, mark size=2, mark options={solid}]
table {%
0.1 0.983601731841868
0.2 2.20144207341122
0.3 2.81385167651191
0.4 3.88497551041231
0.5 4.68604493789953
0.6 5.61678525511839
0.7 6.57718364961296
0.8 7.5605942705256
0.9 8.5256732617083
1 9.4998458078403
1.1 10.4192121995462
1.2 11.3275715083991
1.3 12.2710265299
1.4 13.2437451901419
1.5 14.1200885694931
};

\addplot [semithick, color1, mark=pentagon*, mark size=2, mark options={solid}]
table {%
0.1 0.990069982332072
0.2 2.25990395704983
0.3 2.70872333091596
0.4 3.80501064926266
0.5 4.50913156202263
0.6 5.4149050530622
0.7 6.20886109067688
0.8 7.12383862924983
0.9 8.02109604430928
1 8.82462052819206
1.1 9.62850615973285
1.2 10.456784630145
1.3 11.2553029709396
1.4 12.0960371742056
1.5 12.9337393166587
};

\addplot [semithick, color2, mark=triangle*, mark size=2, mark options={solid}]
table {%
0.1 1.14350352660827
0.2 2.28522598467307
0.3 2.54436090283265
0.4 3.72520520129454
0.5 4.29706779614269
0.6 5.36821819730913
0.7 6.06190909456729
0.8 6.9915603941501
0.9 7.75850036427323
1 8.57578946944022
1.1 9.34231799266898
1.2 10.1130336249478
1.3 10.8030050041086
1.4 11.5804391855254
1.5 12.3022119453597
};

\addplot [semithick, color3, mark=*, mark size=2, mark options={solid}]
table {%
0.1 0.784389965119686
0.2 1.56166142493128
0.3 2.02380271979771
0.4 2.76449909366073
0.5 3.44024892182313
0.6 4.21713865229691
0.7 4.95674122607299
0.8 5.75450186356652
0.9 6.54917073903456
1 7.30744150215547
1.1 8.10752423222077
1.2 8.91167292668826
1.3 9.70651965040006
1.4 10.4844126372556
1.5 11.2733032033494
};

\end{axis}

\end{tikzpicture}\\
        \caption{}
        \label{fig:is_full}
    \end{subfigure} 
    \caption{\small Our model outperforms baseline approaches across the MSE, Dynamic MSE, and IS metrics over the prediction horizon. Lower is better. Note that the per grid cell standard error is too small to be visible in the MSE plots.
    }
    \label{fig:mse_plots}
\end{figure*}
\begin{table*}[t!]
    \centering
    \caption{\small Our model outperforms all baselines across the MSE, Dynamic MSE, and IS metrics on OGM predictions. Lower is better.}
    \begin{tabular}{lcccc}
        \toprule
        \multicolumn{1}{c}{\bf Models} &
        \multicolumn{1}{c}{\bf Parameters $\times 10^{6}$} &
        \multicolumn{1}{c}{\bf MSE $\times 10^{-2}$} &
        \multicolumn{1}{c}{\bf Dynamic MSE$\times 10^{-3}$} &
        \multicolumn{1}{c}{\bf IS}\\
        \midrule
        Ours  & $1.8$ & $\mathbf{3.17 \pm 0.0007}$ & $\mathbf{2.03 \pm 0.0002}$ & $\mathbf{5.86 \pm 0.058}$\\
        MIM~\cite{mim} & $6.8$ & $3.41 \pm 0.0007$ &  $2.23\pm 0.0002$ & $6.86 \pm 0.066$  \\
        PredRNN++~\cite{predrnn++} & $1.8$ & $3.45 \pm 0.0008$ & $2.38\pm 0.0002$ & $7.08 \pm 0.070$\\
        PredNet~\cite{PredNet} & $1.2$ & $3.50\pm 0.0008$  & $2.53\pm 0.0003$ & $7.58 \pm 0.074$\\
        \bottomrule
    \end{tabular}
    \label{table:metrics}
\vspace{-1em}
\end{table*}
\cref{fig:quali_turning} is an example scene with multiple vehicles moving through an intersection. We first consider the vehicle that is making a left turn (circled in the OGMs). Our model is able to retain the vehicle and predict its accurate position, but is not able to predict its turning orientation. We hypothesize that due to the straight motion of the vehicle in the input frames, our model predicts that the vehicle would keep going straight. The ability to predict the turning motion would require modeling the multimodality in the predictions to incorporate many possible vehicle trajectories, which is beyond the scope of our paper, but is a promising avenue for future work. One such possible trajectory is that the vehicle continues to drive straight, which is the prediction of our model. In comparison, the turning vehicle vanishes in the predictions of all other baseline models. Our model is able to capture the structure and relative motion of both of the remaining vehicles in the intersection and the static environment for longer prediction time horizons as compared to the baseline methods. 
Thus, our qualitative results demonstrate that our proposed double-prong architecture successfully learns to predict the relative motion of both the vehicles and the static environment in cluttered driving scenarios.

\subsection{Quantitative Results}
To quantitatively evaluate the prediction performance of our model, we use the per grid cell mean squared error (MSE) and the per frame image similarity (IS)~\cite{im_sim}\footnote{We use this IS implementation:\newline\texttt{https://github.com/BenQLange/Occupancy-Grid-Metrics}.} metrics between the predicted and the target OGMs. The MSE metric is used to measure how well each predicted cell's occupancy probability corresponds to its ground truth value. The IS~\cite{im_sim} metric is used to measure how well the structure of the scene is maintained in the predictions.

\cref{fig:mse_full} shows the plot of the MSE versus the prediction time step for all models up to \SI{1.5}{\second} ahead. The MSE values increase with the prediction time horizon, as expected due to the accumulation of prediction errors. Our model has the lowest MSE for all time steps.
The improvement in model performance becomes very apparent when we isolate the moving object predictions. We compute the MSE for the grid cells corresponding to the dynamic objects, termed dynamic MSE, at each prediction time step, as shown in \cref{fig:mse_dyn}. We apply the same dynamic masks that we use to split the eOGM data to the target and predicted OGMs, then calculate the MSE. 
Our model outperforms the baseline methods in dynamic object predictions, exhibiting consistent performance with the results observed in \cref{fig:qualitative}.

\cref{fig:is_full} shows the plot of the IS metric versus time step on the complete OGM predictions. IS considers the Manhattan distance between the two closest cells of the same class (among occupied, free, and occluded classes) between the target OGMs and the predictions~\cite{im_sim}. Therefore, the lower the IS metric, the better the predictions. \cref{fig:is_full} shows that our model has the lowest IS for all prediction time horizons outside of standard error, indicating that it more accurately predicts the structure of the scenes than baseline techniques.

\cref{table:metrics} shows the averaged MSE, dynamic MSE, and IS metrics over the $15$ prediction time steps. Our model performs better than the baseline methods across all metrics. The MIM~\cite{mim} network generally performs better than the other two baselines, while having the highest number of model parameters. Our model outperforms the MIM~\cite{mim}, PredRNN++~\cite{predrnn++}, and PredNet~\cite{PredNet} baseline methods in MSE by approximately $7.0$\%, $8.7$\%, and $9.4$\%, respectively. We posit that the significant performance improvement achieved by our proposed, parameter-efficient double-prong architecture can be attributed to learning the predictions for the globally static and moving parts of the environment separately. Thus, we are able to effectively incorporate the dynamic scene information directly into our prediction model.

\section{CONCLUSIONS}
\label{conclusions}
We present a double-prong deep neural network architecture for local environment prediction. 
Our model successfully incorporates the environment dynamics directly within its architecture to improve prediction performance on real-world data.
The proposed model is able to accurately predict the motion of sparse dynamic objects for longer prediction time horizons, thus, reducing predicted object disappearance, as compared to baselines methods. Our model is also able to maintain the structure of the environment through egocentric motion.
Since the main assumption in our work is that we have access to tracking information, for future work, we can extend our model to include both object detection and tracking capabilities. Another promising avenue for future work is modeling the multimodality in the predictions.    

\section*{ACKNOWLEDGMENT}
This project was made possible by funding from the Ford-Stanford Alliance. 
We thank Dr. Marcos Paul Gerardo Castro for his advice and assistance. We thank Bernard Lange for help with implementing the PredRNN++ baseline architecture and image similarity metric.  

\renewcommand*{\bibfont}{\small}
\printbibliography
\end{document}